\documentclass{article}

\usepackage[nonatbib]{nips_2017}

\usepackage{nips_2017}

\usepackage[utf8]{inputenc} 
\usepackage[T1]{fontenc}    
\usepackage{hyperref}       
\usepackage{url}            
\usepackage{booktabs}       
\usepackage{amsfonts}       
\usepackage{nicefrac}       
\usepackage{microtype}      
\usepackage{amsmath}
\usepackage{graphicx}
\usepackage{wrapfig}
\usepackage{caption,setspace}
\usepackage{ mathrsfs }
\usepackage{subcaption}
\usepackage{adjustbox} 

\usepackage{color}

\makeatletter
\newcommand*{\centerfloat}{%
  \parindent \z@
  \leftskip \z@ \@plus 1fil \@minus \textwidth
  \rightskip\leftskip
  \parfillskip \z@skip}
\makeatother
\title{An Improved Training Procedure for Neural Autoregressive Data Completion}

\author{
  Maxime Voisin \\
  Stanford University\\
  \texttt{maximev@cs.stanford.edu} \\
  \And
  Daniel Ritchie \\
  Brown University \\
  \texttt{daniel\_ritchie@brown.edu} \\
}

\begin{document}

\maketitle

\section{Introduction}

Neural autoregressive models are explicit density estimators that achieve state-of-the-art likelihoods for generative modeling \cite{oord2016pixel,salimans2017pixelcnn++,uria2016neural,van2016conditional}. The D-dimensional data distribution $p(x)$ is factorized into an autoregressive product of one-dimensional conditional distributions according to the chain rule. Each conditional distribution is parametrized by a shared neural network.

Data completion is a more involved task than data generation: the model must infer missing variables for any partially observed input vector. Previous work \cite{uria2016neural} introduced an order-agnostic training procedure for data completion with autoregressive models. It maximizes the average likelihood of the model over all $D!$ orderings of the data dimensions. As a result, all possible one-dimensional conditionals $p(x_{o_d} | x_{o_1} 
,...,x_{o_{d-1}})$ are trained, for any $d \in \{1..D\}$ and any ordering $o$ of $\{1...n\}$. Thus, missing variables in any partially observed input vector can be imputed efficiently by choosing an ordering where observed dimensions precede unobserved ones and by computing the autoregressive product in this order. 
This training procedure can be made efficient: \cite{uria2014deep} estimates the order-agnostic loss with an unbiased estimator that reuses most computations.

In this paper, we provide evidence that the order-agnostic (OA) training procedure is suboptimal for data completion. 
We propose an alternative procedure (OA++) that reaches better performance in fewer computations. It can handle all data completion queries while training fewer one-dimensional conditional distributions than the OA procedure. In addition, these one-dimensional conditional distributions are trained proportionally to their expected usage at inference time, reducing overfitting. Finally, our OA++ procedure can exploit prior knowledge about the distribution of inference completion queries, as opposed to OA. We support these claims with quantitative experiments on standard datasets used to evaluate autoregressive generative models.

\section{Improving the order-agnostic loss for data completion}
The OA procedure trains autoregressive models for data completion by optimizing the loss $\mathscr{I}_{\text{OA}}(\theta)$ in Equation \ref{eq:1}. In practice, the exact loss has too many terms to be computationally tractable. It is estimated by sampling a training vector $x$ uniformly at random, a number $d \in \{1..D\}$ and a set $o_{<d}$ of conditioned variables uniformly at random, and by computing $\widehat{\mathscr{I}}_{\text{OA}}(\theta)$ as in Equation \ref{eq:2}. The sum in $\widehat{\mathscr{I}}_{\text{OA}}(\theta)$ is computed over all possible choices of $o_d$, and the neural network computations involved in $p(x_{o_d} | x_{o<d}; \theta,o_d)$ can be reused across the terms of the sum.

\begin{align}
\centerfloat
\label{eq:1}
& \mathscr{I}_{\text{OA}}(\theta) = \mathbb{E}_{x \in X} \mathbb{E}_{o \in D!} -\log p(x ; \theta, o) = \mathbb{E}_{x \in X} \mathbb{E}_{o \in D!}  - \sum_{d=1}^{D} \log p(x_{o_d} | x_{o<d} ;\theta, o) \\
\label{eq:2}
& \approx \widehat{\mathscr{I}}_{\text{OA}}(\theta) = \frac{D}{D-d+1} \sum_{o_d} - \log p(x_{o_d} | x_{o<d}; \theta, o_{<d},o_d)
\end{align} 

In this section, we present and motivate step-by-step the modifications we propose to the OA procedure (Equation \ref{eq:1}), leading to the OA++ procedure (Equation \ref{eq:4}).

\subsection{Equal use of 1D conditional distributions at training and inference}
Ideally, the usage of each 1D conditional distribution at inference time should be proportional to its usage at training time. 
Otherwise, some 1D conditional distributions frequently used at inference time might be under-trained (undermining performance), and some 1D conditional distributions rarely used at inference time might be over-trained (undermining efficiency). For $d\in \{1..D\}$, we refer to 1D conditional distributions  with $d-1$ conditioned variables, $p(x_{o_d} | x_{o<d}; \theta)$, as 1D conditional distributions of size $d$.

\textbf{Result}: Under the OA procedure, at any training iteration, each 1D conditional distribution of size $d$  has a fixed probability $\frac{1}{\binom{D}{d-1}(D-d+1)}$ of being trained.

\textit{Proof}: At any training iteration, exactly one 1D conditional distribution of size $d$ is trained. It is chosen uniformly at random among the $\binom{D}{d-1}(D-d+1)$ 1D conditional distributions of size $d$.

\textbf{Result}: Under the OA procedure, each 1D conditional distribution of size $d$ has probability $\frac{1}{\binom{D}{d-1}(D-d+1)}$ of being involved in any \emph{generation} query.

\textit{Proof}: Exactly one 1D conditional distribution of size $d$ is involved in any \textit{generation} query, as each of the $D$ variables must be sampled in sequence. There are  $\binom{D}{d-1}(D-d+1)$ such distributions.
The OA procedure chooses a variable ordering uniformly at random.
Thus, any 1D conditional distribution of size $d$ has an equal probability $\frac{1}{\binom{D}{d-1}(D-d+1)}$ of being used in any \textit{generation} query.

Therefore, under the OA procedure, the expected usage of any 1D conditional distribution at inference time is proportional to its expected usage at training time for \emph{generation} queries.
We are interested in data \textit{completion} queries, however. 
In a completion query, the values of some variables are known, and thus conditional distributions for those variables do not need to be computed.
This changes the expected usage patterns for each size $d$ of conditional distribution. 
Using the OA procedure leads to a discrepancy between the usage of 1D conditional distributions during training and inference.
Further, if we know something in advance about the distribution of inference queries (e.g. which variables will be known, or how many will be known), the OA procedure has no way to exploit such prior knowledge.

The OA++ procedure we propose does not suffer from these limitations.
It assumes that inference queries follow a distribution $\mathscr{D}$, and it trains all 1D conditional distributions proportionally to their expected usage during inference under this distribution.
If we have prior knowledge about the expected structure of inference queries, this can be encoded in $\mathscr{D}$.
If we have no such prior knowledge, then OA++ sets $\mathscr{D}$ to be a uniform distribution over inference queries, i.e. the set of observed variables in a completion query is drawn uniformly at random.

Instead of optimizing $\log p(x;\theta,o)$ over all D! orderings of $\{1..n\}$ as in the OA procedure,
OA++ samples  $ \textbf{obs} \subset \{1..D\}$ from the expected distribution of inference queries $\mathscr{D}$, it samples an ordering $o$ of the unobserved input $x^\textbf{missing}$ uniform at random, and it optimizes $\log p(x^\textbf{missing} | x^\textbf{obs} ;\theta,o)$. Since $\textbf{missing} \cup \textbf{obs} = \{1..D\}$, there are $(D-|\textbf{obs}|)!$ possible orderings $o$ of the unobserved input. The corresponding loss is expressed in Equation \ref{eq:3}: 
\begin{equation}
\label{eq:3}
\mathbb{E}_{x \in X}  \mathbb{E}_{\textbf{obs} \sim \mathscr{D} } \mathbb{E}_{o \in (D-|\textbf{obs}|)!} -\log p(x^{\textbf{missing}} | x^\textbf{obs} ; \theta, o) \\
\end{equation}

\subsection{Training fewer 1D conditional distributions}

The OA procedure maximizes the average log-likelihood of the model over all orderings: in Equation \ref{eq:1}, the ordering $o$ is treated as a uniform random variable over its $D!$ values. Consequently, all 1D conditional distributions are trained. 
However, these conditional distributions will not all be used.
To handle any given query, the model must either fix an order of the unobserved variables, or use an ensemble of $K>1$ orderings of the unobserved variables as in \cite{uria2016neural}.
In most settings, $K << D!$, thus far fewer 1D conditionals will be used than were trained.
Since the parameters of all conditionals are determined by a single shared neural network, the model is wasting its representational capacity on 1D conditionals that will not be used at inference time.

The OA++ procedure we propose instead trains at most $K \sum_{d=1}^D \binom{D}{d-1}$ conditional distributions, much fewer than the total number $\sum_{d=1}^D \binom{D}{d-1}(D-d+1) $ of 1D conditional distributions, and it can still handle any completion query by using an ensemble of K orderings.
To do so, we fix in advance K orderings $O_K = \{o_1,...,o_K\}$ of $\{1..D\}$. For any completion query $x^\textbf{obs}$, an ordering $o$ is sampled uniform at random from $O_K$. The autoregressive sum is computed over the unobserved input $x^\textbf{missing}$ (data dimensions are ordered according to $o$). Fundamentally, OA++ treats the ordering $o$ as a uniform random variable over K values, instead of $D!$ values for OA.

The OA++ loss $\mathscr{I}_{\text{OA++}}(\theta)$ for data completion is expressed as in Equation \ref{eq:4}. $\mathscr{D}$  is the distribution of inference queries if we have prior knowledge of it; otherwise $\mathscr{D}$ is the uniform random distribution of completion queries. K is the maximum number of inference queries that we expect to average by ensembling at inference time. For a given completion query $\textbf{obs} \in \mathscr{D}$, the $D-|\textbf{obs}|$ data dimensions of $x^{\textbf{missing}}$ are reordered according to ordering $o$.
\begin{align}
\mathscr{I}_{\text{OA++}}(\theta) &= \mathbb{E}_{x \in X} \mathbb{E}_{o \in O_K}  \mathbb{E}_{\textbf{obs} \sim \mathscr{D} } -\log p(x^{\textbf{missing}} | x^\textbf{obs} ; \theta, o) \\
\label{eq:4}
&= \mathbb{E}_{x \in X} \mathbb{E}_{o \in O_K}  \mathbb{E}_{\textbf{obs} \sim \mathscr{D} } \sum_{d=1}^{D-|\textbf{obs}|} - \log p(x^{\textbf{missing}}_{d} | x^\textbf{obs}, x^{\textbf{missing}}_{<d} ; \theta, o)
\end{align}

\textbf{Unbiased Estimator}
The OA++ loss in \ref{eq:4} has a large number of terms. It can be estimated by sampling a training example $x$ uniformly at random, a completion query $\textbf{obs} \sim \mathscr{D}$, a number $d \in \{1..D-|obs|\}$, and by computing $\widehat{\mathscr{I}}_{\text{OA++}}(\theta)$ as in Equation \ref{eq:6}.   
\begin{equation}
\label{eq:6}
\widehat{\mathscr{I}}_{\text{OA++}}(\theta) = - (D-|\textbf{obs}|) \log p(x^{\textbf{missing}}_{d} | x^\textbf{obs}, x^{\textbf{missing}}_{<d} ; \theta, o)
\end{equation}
It is possible to provide an unbiased estimator of the loss, that consists of a sum with computations shared among its different terms, like in Equation \ref{eq:2}. However, under the assumption $K<<D!$, the sum would likely consist  of a single term, effectively reducing to Equation \ref{eq:6}.

\textbf{Unifying data generation and data completion}\\
By choosing $K=1$, OA++ reduces to the original training procedure for NADEs introduced in \cite{larochelle2011neural}.
By choosing $K=D!$, OA++ reduces to the order-agnostic training procedure for data \textit{generation} introduced in \cite{uria2014deep}. In other words, OA++ unifies the training procedures of data generation and data completion under a single framework.

\section{Results}
In order to compare OA and OA++, we train the same autoregressive model (two-layer NADE) with both procedures.
We conduct experiments on eight multivariate binary datasets commonly used by previous works on autoregressive models \cite{germain2015made,larochelle2011neural,uria2016neural}. Table \ref{table:full_loss_results} reports the performance of all models on 2 test sets of inference queries: one consists of uniform random completion queries; the other consists of completion queries of size $D/2$ picked at random. On the second test set, models were provided prior knowledge of the distribution of inference queries. Results are computed for $K=1$ (no ensemble learning). The comparison of OA and OA++ for $K>1$ is left as future work. 

\textbf{Performance at convergence}
OA++ outperforms OA on all experiments and does especially well when given prior knowledge about the distribution of inference queries.

\begin{table*}
\centering
\resizebox{\columnwidth}{!}{%
\begin{tabular}{lcccccccc}
\toprule   
  Model  & Adult & DNA & Mushrooms & NIPS-0-12 &  Connect4 & OCR-letters & RCV1 & Web\\ 
\midrule
OA & 9.8 / 13.6 & 87.2 / 90.7  & 5.2 / 9.5 & 277.0 / 280.7 & 9.4 / 14.7 & 31.7 37.8 & 47.0 / 48.1 & 28.8 / 30.1\\
\midrule
OA++ & \textbf{7.8} / \textbf{11.9} & \textbf{78.2} / \textbf{83.3}  & \textbf{4.2} / \textbf{7.8} & \textbf{272.0} / \textbf{276.4} & \textbf{4.7} / \textbf{9.5} & \textbf{23.0} / \textbf{27.7} & \textbf{46.0} / \textbf{47.2} & \textbf{27.9} / \textbf{29.1}\\
\bottomrule\\

\end{tabular}
}
\caption{ Average negative log-likelihood of NADE models on 2 test sets of  inference queries. In each cell, the left metric corresponds to uniform random queries, the right metric corresponds to random queries of size D/2 known in advance} 
\label{table:full_loss_results} 
\end{table*}

\begin{figure}
\centerfloat
\begin{tabular}{cccc}
\includegraphics[width = 1.8in]{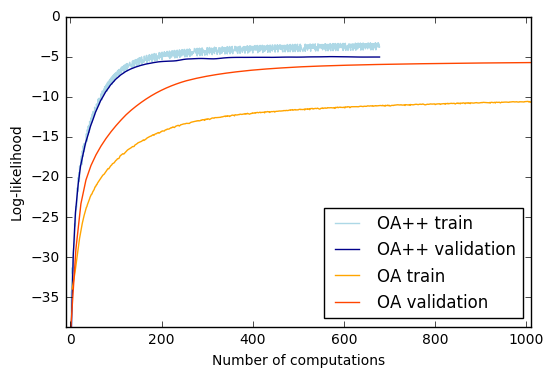} 
{\includegraphics[width = 1.8in]{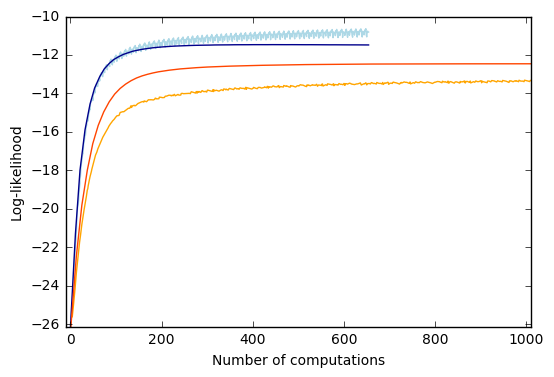} }\\

{\includegraphics[width = 1.8in]{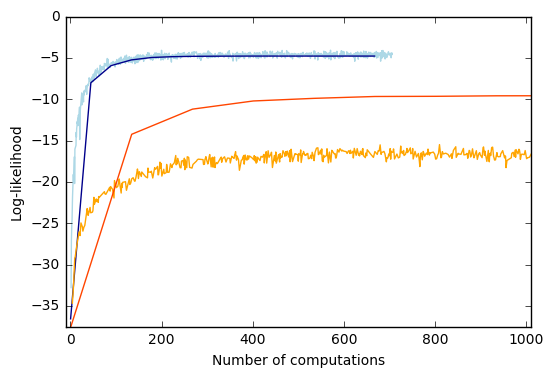}} 
{\includegraphics[width = 1.8in]{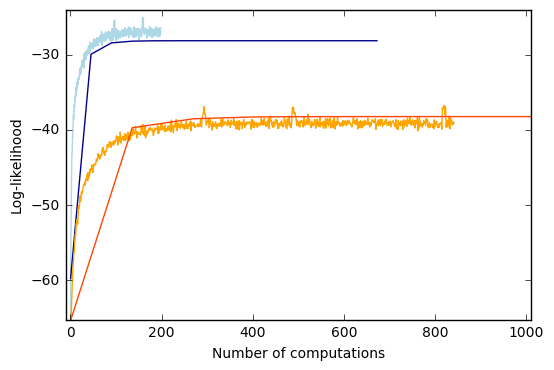}} 
\end{tabular}
\caption{Evolution of the train and validation loglikelihoods with the number of computations. Computations are reported as the number of neural network inferences performed, divided by the dimensionality D of the data. From left to right and from top to bottom: Mushrooms, Adult, Connect4, OCRletters. }
\label{3}
\end{figure}

\textbf{Speed of convergence} 
OA++ converges in fewer computations than OA. OA++ also does not suffer from overfitting, while OA sometimes does. Figure \ref{3} reports the evolution of the training and validation log-likelihoods with the number of computations, for some experiments.

\end{document}